\documentclass{article} 
\usepackage{iclr2023_conference_tinypaper,times}
\pdfoutput=1

\usepackage{amsmath,amsfonts,bm}

\def\eqref#1{equation~\ref{#1}}

\def\1{\bm{1}}

\DeclareMathAlphabet{\mathsfit}{\encodingdefault}{\sfdefault}{m}{sl}
\SetMathAlphabet{\mathsfit}{bold}{\encodingdefault}{\sfdefault}{bx}{n}

\usepackage[inline]{enumitem} 
\usepackage{color}
\usepackage{multirow}
\usepackage[makeroom]{cancel}
\usepackage[accsupp]{axessibility}

\usepackage{blindtext}

\definecolor{turquoise}{cmyk}{0.65,0,0.1,0.3}
\definecolor{purple}{rgb}{0.65,0,0.65}
\definecolor{dark_purple}{rgb}{0.5,0,0.5}
\definecolor{dark_green}{rgb}{0, 0.5, 0}
\definecolor{orange}{rgb}{0.8, 0.6, 0.2}
\definecolor{red}{rgb}{0.8, 0.2, 0.2}
\definecolor{darkred}{rgb}{0.6, 0.1, 0.05}
\definecolor{blueish}{rgb}{0.0, 0.3, .6}
\definecolor{light_gray}{rgb}{0.7, 0.7, .7}
\definecolor{pink}{rgb}{1, 0, 1}
\definecolor{greyblue}{rgb}{0.25, 0.25, 1}

\newcommand{\revise}[1]{{\color{black}#1}}

\usepackage{hyperref}
\usepackage{url}
\usepackage{graphicx}
\usepackage{wrapfig}

\title{Exploiting GPT-4 Vision for \\
Zero-shot Point Cloud Understanding}

\iclrfinalcopy
\author{Qi Sun\footnotemark[1]
, Xiao Cui\footnotemark[1], Wengang Zhou\footnotemark[2], Houqiang Li\footnotemark[2] \\
Department of EEIS, University of Science and Technology of China \\
\texttt{\{qisun, cuixiao2001\}@mail.ustc.edu.cn,\{zhwg, lihq\}@ustc.edu.cn}
}

{
  \renewcommand{\thefootnote}%
    {\fnsymbol{footnote}}
  \footnotetext[1]{Equal contribution.}
  \footnotetext[2]{Corresponding Authors.}
}

\begin{document}

\maketitle

\begin{abstract}
In this study, we tackle the challenge of classifying the object category in point clouds, which previous works like PointCLIP struggle to address due to the inherent limitations of the CLIP architecture. Our approach leverages GPT-4 Vision (GPT-4V) to overcome these challenges by employing its advanced generative abilities, enabling a more adaptive and robust classification process. 
\revise{We adapt the application of GPT-4V to process complex 3D data, enabling it to achieve zero-shot recognition capabilities without altering the underlying model architecture.}
Our methodology also includes a systematic strategy for point cloud image visualization, mitigating domain gap and enhancing GPT-4V's efficiency. Experimental validation demonstrates our approach's superiority in diverse scenarios, setting a new benchmark in zero-shot point cloud classification.
\end{abstract}

\section{Introduction}
\begin{figure}[b]
    \centering
    \includegraphics[width=\linewidth]{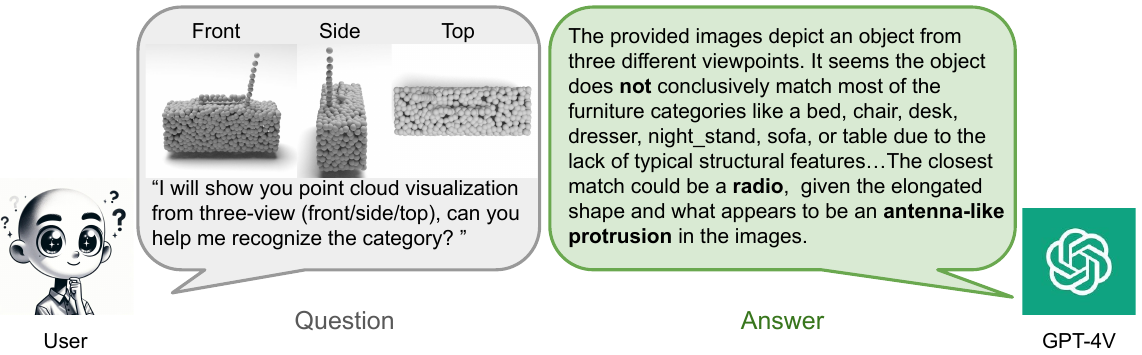}
    \caption{Illustration for our method. Using three-view point cloud rendered images and a predefined text template as input, GPT-4V will analyse the visual clue like human then point out the category.}
    \label{fig:framework}
\end{figure}

3D point cloud understanding has many applications in autonomous driving, robotics and scene understanding. 
Point-based methods~\citep{qi2017pointnet, qi2017pointnet++, liu2019relation, zhang2023uni3d, yu2022point, qi2023contrast, cheraghian2022zero} learn features directly from raw point cloud, while projection-based methods ~\citep{goyal2021revisiting, sarkar2018learning, roveri2018network} learn the projected 2D image features. 
Recent research efforts have been directed towards zero-shot understanding of point clouds~\citep{zhu2022pointclip, zhang2022pointclip, huang2023clip2point}, employing models pretrained exclusively on 2D images. Nevertheless, the effectiveness of these approaches is inherently limited by the characteristics of CLIP, due to its contrastive training strategy and the domain discrepancy between the visualizations of point clouds and the associated textual labels.


Our methodology addresses this challenge by leveraging GPT-4 Vision~\citep{openai2023gpt4}
Utilizing GPT-4V's advanced generative power
our approach transcends the constraints of similarity-based classification. 
With profound and integrative analysis of both text and images,
it can adapt effectively to various image formats through tailored prompt templates. Also, it has great interpretability capabilities. Instead of merely providing an choice, it explicitly indicates the specific attributes or features that inform its decision-making process, which mirrors aspects of human cognition.





While GPT-4 exhibits enhanced capabilities in aligning text and images, its performance efficiency still fluctuates with different point cloud visualization methods. Our study identifies the most effective visualization technique to maximize GPT-4's potential and provides a detailed discussion of the underlying reasons, paving the way for future works.

\section{Methods}
\begin{table}[]
    \centering
    \caption{\textbf{Quantitative results:} comparison with the-state-of-the-art methods on classification accuracy ($\%$). $(K)$ means $K$-view images used for classification. ``$\star$'' denotes that  GPT-4V sometimes encounters error when generating responses, in which cases we don't account for accuracy.}
    \resizebox{\linewidth}{!}
    {\begin{tabular}{l|cccccc}
        Datasets  & PointCLIP (3) & PointCLIP (6) & PointCLIP V2 (3) & PointCLIP V2 (10) & Ours (3) \\
        \hline
        ModelNet10 & 16.0 & 24.0 & 44.0 & 66.0 & \textbf{72.7$^{\star}$} (32/44)\\
        ModelNet40 & 12.0 & 12.0 & 50.0 & 56.0 & \textbf{58.7$^{\star}$} (28/46)
    \end{tabular}}
    \label{tab:pc}
\end{table}
\begin{table}[h]
    \centering
    \caption{\textbf{Ablation study:} different visualization methods in ModelNet10 dataset. 
    ``DM'' / ``RI'' represents depth map / rendered image, respectively.}
    \resizebox{\linewidth}{!}{
    \begin{tabular}{l|cccccc}
        \revise{Visualizations}  &  DM-sparse (3) & DM-dense (3) & RI-colored (3) & RI-gray  (1) & RI-gray  (3) \\
        \hline
        Accuracy ($\%$) & 13.3$^{\star}$ (6/45) & 70.7$^{\star}$ (29/41)& 52.2$^{\star}$ (24/46) & 64.0 & \textbf{72.7}$^{\star}$ (32/44)
    \end{tabular}}
    \vspace{-5mm}
    \label{tab:abl}
\end{table}


\label{Methods}
The task of point cloud classification can be formulated as a mapping $f: x \in \mathbb{R}^{K\times3} \rightarrow l \in \mathbb{R}^{C}$, where $K$ is the number of unoriented point cloud and $C$ is the category numbers.
The method is straightforward: we input the visualized point cloud and predefined question template with the category options to GPT-4V, then ask it to give us the object class.
To harness the visual-linguistic comprehension abilities of GPT-4V, we first employ various visualization methods to convert the 3D point cloud to RGB images $I \in \mathbb{R}^{H\times W \times3}$.  
To mitigate information loss from 3D-to-2D projection, we use three distinct views (side/front/top) that are widely adopted in CAD engineering. As depicted in Figure~\ref{fig:framework}, GPT-4V will conduct visual analysis to identify and determine the object category.
The specifics of these prompt templates and the visualization methods are comprehensively detailed in the Appendix~\ref{sec:prompts} for the reproducibility.


\section{Experiments}
\label{exp}

\paragraph{Settings.} In our experiments, we utilize two datasets: ModelNet10~\citep{wu20153d} and ModelNet40~\citep{wu20153d}. Due to the constraints imposed by the GPT-4V web service, we are limited to selecting 50 point cloud samples from the original validation dataset.
As baselines, we choose two of the most representative zero-shot point cloud classification methods based on CLIP~\citep{radford2021learning}: PointCLIP~\citep{zhang2022pointclip} and PointCLIP V2~\citep{zhu2022pointclip}.  

\paragraph{Results.} 
The quantitative results are presented in Table~\ref{tab:pc}, where our method demonstrates state-of-the-art performance on both datasets. Notably, our approach outperforms the second-best method by a substantial margin of 6.7$\%$ on ModelNet10. 
It's worth mentioning that our approach utilizes only three views to represent a single point cloud, which is significantly fewer compared to the requirements of PointCLIP and PointCLIP V2.

\paragraph{Discussions.} 
In Table~\ref{tab:abl}, we present various visualization methods, including sparse depth map, dense depth map, colored rendered image and gray rendered image, for point cloud three-views input to GPT-4V, and their significant impact on classification accuracy becomes evident. Details on visualization are provided in Appendix~\ref{app:vis}.
Among the visualization techniques, the grayscale rendering can effctively convey the shape and distinctive features of the point cloud. On the other hand, depth maps, whether sparse or dense, yield lower-resolution projections that fail to capture precise geometry adequately.
When using colored rendering, there is a potential for misunderstanding by GPT-4V due to the presence of colored textures. It's important to note that we provided prompts indicating that the colors merely represent different point locations.
Furthermore, our experiments reveal that employing multi-views aids GPT-4V in recognizing point cloud categories. In contrast, a single-view setting results in a noticeable performance drop of 8.7$\%$.


\subsubsection*{URM Statement}
The authors acknowledge that at least one key author of this work meets the URM criteria of ICLR 2024 Tiny Papers Track.

\bibliography{iclr2023_conference_tinypaper}

\begin{thebibliography}{17}
\providecommand{\natexlab}[1]{#1}
\providecommand{\url}[1]{\texttt{#1}}
\expandafter\ifx\csname urlstyle\endcsname\relax
  \providecommand{\doi}[1]{doi: #1}\else
  \providecommand{\doi}{doi: \begingroup \urlstyle{rm}\Url}\fi

\bibitem[Cheraghian et~al.(2022)Cheraghian, Rahman, Chowdhury, Campbell, and Petersson]{cheraghian2022zero}
Ali Cheraghian, Shafin Rahman, Townim~F Chowdhury, Dylan Campbell, and Lars Petersson.
\newblock Zero-shot learning on {3D} point cloud objects and beyond.
\newblock \emph{IJCV}, 2022.

\bibitem[Goyal et~al.(2021)Goyal, Law, Liu, Newell, and Deng]{goyal2021revisiting}
Ankit Goyal, Hei Law, Bowei Liu, Alejandro Newell, and Jia Deng.
\newblock Revisiting point cloud shape classification with a simple and effective baseline.
\newblock In \emph{ICML}, 2021.

\bibitem[Huang et~al.(2023)Huang, Dong, Yang, Huang, Lau, Ouyang, and Zuo]{huang2023clip2point}
Tianyu Huang, Bowen Dong, Yunhan Yang, Xiaoshui Huang, Rynson~WH Lau, Wanli Ouyang, and Wangmeng Zuo.
\newblock {CLIP2point}: Transfer {CLIP} to point cloud classification with image-depth pre-training.
\newblock In \emph{ICCV}, 2023.

\bibitem[Liu et~al.(2019)Liu, Fan, Xiang, and Pan]{liu2019relation}
Yongcheng Liu, Bin Fan, Shiming Xiang, and Chunhong Pan.
\newblock Relation-shape convolutional neural network for point cloud analysis.
\newblock In \emph{CVPR}, 2019.

\bibitem[OpenAI(2023)]{openai2023gpt4}
OpenAI.
\newblock {GPT}-4 technical report.
\newblock \emph{arXiv preprint arXiv:2303.08774}, 2023.

\bibitem[Qi et~al.(2017{\natexlab{a}})Qi, Su, Mo, and Guibas]{qi2017pointnet}
Charles~R Qi, Hao Su, Kaichun Mo, and Leonidas~J Guibas.
\newblock {PointNet}: Deep learning on point sets for {3D} classification and segmentation.
\newblock In \emph{CVPR}, 2017{\natexlab{a}}.

\bibitem[Qi et~al.(2017{\natexlab{b}})Qi, Yi, Su, and Guibas]{qi2017pointnet++}
Charles~Ruizhongtai Qi, Li~Yi, Hao Su, and Leonidas~J Guibas.
\newblock {PointNet}++: Deep hierarchical feature learning on point sets in a metric space.
\newblock \emph{NeurIPS}, 2017{\natexlab{b}}.

\bibitem[Qi et~al.(2023)Qi, Dong, Fan, Ge, Zhang, Ma, and Yi]{qi2023contrast}
Zekun Qi, Runpei Dong, Guofan Fan, Zheng Ge, Xiangyu Zhang, Kaisheng Ma, and Li~Yi.
\newblock Contrast with reconstruct: Contrastive {3D} representation learning guided by generative pretraining.
\newblock In \emph{ICML}, 2023.

\bibitem[Radford et~al.(2021)Radford, Kim, Hallacy, Ramesh, Goh, Agarwal, Sastry, Askell, Mishkin, Clark, et~al.]{radford2021learning}
Alec Radford, Jong~Wook Kim, Chris Hallacy, Aditya Ramesh, Gabriel Goh, Sandhini Agarwal, Girish Sastry, Amanda Askell, Pamela Mishkin, Jack Clark, et~al.
\newblock Learning transferable visual models from natural language supervision.
\newblock In \emph{ICML}, 2021.

\bibitem[Roveri et~al.(2018)Roveri, Rahmann, Oztireli, and Gross]{roveri2018network}
Riccardo Roveri, Lukas Rahmann, Cengiz Oztireli, and Markus Gross.
\newblock A network architecture for point cloud classification via automatic depth images generation.
\newblock In \emph{CVPR}, 2018.

\bibitem[Sarkar et~al.(2018)Sarkar, Hampiholi, Varanasi, and Stricker]{sarkar2018learning}
Kripasindhu Sarkar, Basavaraj Hampiholi, Kiran Varanasi, and Didier Stricker.
\newblock Learning {3D} shapes as multi-layered height-maps using {2D} convolutional networks.
\newblock In \emph{ECCV}, 2018.

\bibitem[Uy et~al.(2019)Uy, Pham, Hua, Nguyen, and Yeung]{uy2019revisiting}
Mikaela~Angelina Uy, Quang-Hieu Pham, Binh-Son Hua, Thanh Nguyen, and Sai-Kit Yeung.
\newblock Revisiting point cloud classification: A new benchmark dataset and classification model on real-world data.
\newblock In \emph{ICCV}, 2019.

\bibitem[Wu et~al.(2015)Wu, Song, Khosla, Yu, Zhang, Tang, and Xiao]{wu20153d}
Zhirong Wu, Shuran Song, Aditya Khosla, Fisher Yu, Linguang Zhang, Xiaoou Tang, and Jianxiong Xiao.
\newblock {3D ShapeNets}: A deep representation for volumetric shapes.
\newblock In \emph{CVPR}, 2015.

\bibitem[Yu et~al.(2022)Yu, Tang, Rao, Huang, Zhou, and Lu]{yu2022point}
Xumin Yu, Lulu Tang, Yongming Rao, Tiejun Huang, Jie Zhou, and Jiwen Lu.
\newblock Point-{BERT}: Pre-training {3D} point cloud transformers with masked point modeling.
\newblock In \emph{CVPR}, 2022.

\bibitem[Zhang et~al.(2023)Zhang, Yuan, Shi, Chen, Li, and Qiao]{zhang2023uni3d}
Bo~Zhang, Jiakang Yuan, Botian Shi, Tao Chen, Yikang Li, and Yu~Qiao.
\newblock {Uni3D}: A unified baseline for multi-dataset {3D} object detection.
\newblock In \emph{CVPR}, 2023.

\bibitem[Zhang et~al.(2022)Zhang, Guo, Zhang, Li, Miao, Cui, Qiao, Gao, and Li]{zhang2022pointclip}
Renrui Zhang, Ziyu Guo, Wei Zhang, Kunchang Li, Xupeng Miao, Bin Cui, Yu~Qiao, Peng Gao, and Hongsheng Li.
\newblock {PointCLIP}: Point cloud understanding by {CLIP}.
\newblock In \emph{CVPR}, 2022.

\bibitem[Zhu et~al.(2023)Zhu, Zhang, He, Zeng, Zhang, and Gao]{zhu2022pointclip}
Xiangyang Zhu, Renrui Zhang, Bowei He, Ziyao Zeng, Shanghang Zhang, and Peng Gao.
\newblock {PointCLIP V2}: Adapting {CLIP} for powerful {3D} open-world learning.
\newblock In \emph{ICCV}, 2023.

\end{thebibliography}
\bibliographystyle{iclr2023_conference_tinypaper}

\appendix
\section{Visualization methods and text prompts}

\label{app:vis}

For the reproducibility of our work, we provide the concrete descriptions on different visualizations and additional text prompts. In addition, data visualization methods, datasets and baselines are provided in the anonymous link\footnote{\url{https://anonymous.4open.science/r/GPT4-V-pointcloud-16FE/}}.

\begin{figure}[h]
    \centering
    \includegraphics[width=\linewidth]{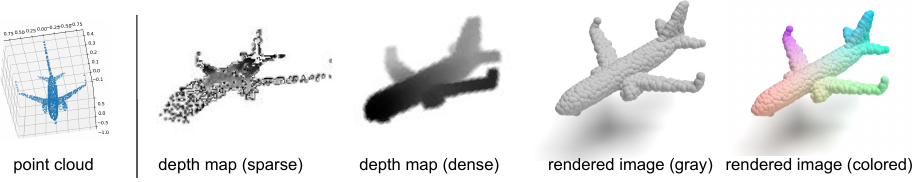}
    \caption{3D point cloud (left) and four different point cloud visualization methods (right). }
    \label{fig:vis}
\end{figure}
\label{sec:prompts}

\paragraph{Rendered image.} 
The code we use for rendering images from point clouds is derived from PointFlowRenderer\footnote{\url{https://github.com/zekunhao1995/PointFlowRenderer}}, which utilizes the physics-based rendering engine Mitsuba (\url{http://www.mitsuba-renderer.org/}) for generating the visuals. 
For the top, front, and side views, we set the camera origins at \texttt{(0,0,3)}, \texttt{(3,0,1)}, and \texttt{(0,-3,1)} respectively, all directed towards the coordinate origin at \texttt{(0,0,0)}.
In rendering the point clouds, two approaches are used: for the gray-scale version, each point is rendered in a uniform gray color, with RGB values set to \texttt{(123/255, 123/255, 123/255)}. In the colored version, the RGB color values for each point \texttt{(r, g, b)} are determined based on the normalized location of the point \texttt{(x, y, z)} in the 3D space.

\paragraph{Depth map (sparse).} 
Initially, we project the 3D point, denoted as \texttt{(x, y, z)}, directly onto the image plane, resulting in foreshortened figures, where the size of the figures varies with depth – smaller for points that are farther away and larger for those that are closer. Following this projection, the resulting value for each point is replicated across three channels to create a three-channel RGB image. Code is built upon the official PointCLIP implementation\footnote{\url{https://github.com/ZrrSkywalker/PointCLIP}}.

\paragraph{Depth map (dense).}
To transform a point cloud into a dense and realistic depth map, we follow a multi-step procedure. The process begins by quantizing the continuous point cloud into sparse voxel grids. Next, we densify these grids using a local mini-value pooling operation, which helps in filling gaps and creating a more continuous spatial representation. Following this, a non-parametric Gaussian kernel is applied for shape smoothing and noise filtering, enhancing the quality of the representation by reducing irregularities and artifacts. Finally, we compress the depth dimension of the voxel grid, resulting in the final projected depth map.
Code is borrowed from the official PointCLIP V2\footnote{\url{https://github.com/yangyangyang127/PointCLIP_V2}}.

\paragraph{Text prompts.}
``I will show you \texttt{\{type of point cloud visualization\}} from three-view (front/side/top) of an object, can you help me recognize the category?  I will provide you \texttt{\{C\}} options: \texttt{\{category list\}}. choose one. Focus on the shape and distinctive features. Please evaluate each possible class respectively." 

Note that \texttt{C} is set to 10/40 for ModelNet10/ModelNet40 dataset, respectively. \texttt{\{type of point cloud visualization\}} can be either \texttt{sparse depth map projected by point cloud} / \texttt{dense depth map projected by point cloud} / \texttt{point cloud visualization}.

\revise{

\section{Ablation study on number of views}
Table~\ref{tab:nov} showcases the classification accuracy in different number of views. Due to GPT-4V API's limitation of accepting a maximum of four images per input, we combine N images (when N $>$ 4) into a single composite image for system input. Consistent with prior work, we typically use either 6 or 10 views. Initially, using six views results in a minor decrease in accuracy, attributable to the different input image form and the absence of explicit viewing angles in the text prompts. However, as we increase the number of views to 10, we observe an improvement in GPT-4V's performance since additional views provide more comprehensive detail and features of the object, addressing many failure cases in 6-views.
\begin{table}[]
    \centering
    \caption{\color{black}Ablation study on number of image views for GPT-4V input. ``$\star$'' denotes that  GPT-4V sometimes encounters error when generating responses, in which cases we don't account for accuracy.}
    \resizebox{0.5\linewidth}{!}
    {
    \color{black} 
    \begin{tabular}{l|cccc}
        Number of views  & 1 & 3  & 6 & 10 \\
        \hline
          Accuracy ($\%$) & 64.0 & 72.7$^\star$ (32/44) & 66.0 & 76.0
        \vspace{-8mm}
    \end{tabular}}
    \label{tab:nov}
\end{table}

}

\section{Case demonstration}
\label{sec:qualitative}
\begin{figure}[h]
    \centering
    \includegraphics[width=\linewidth]{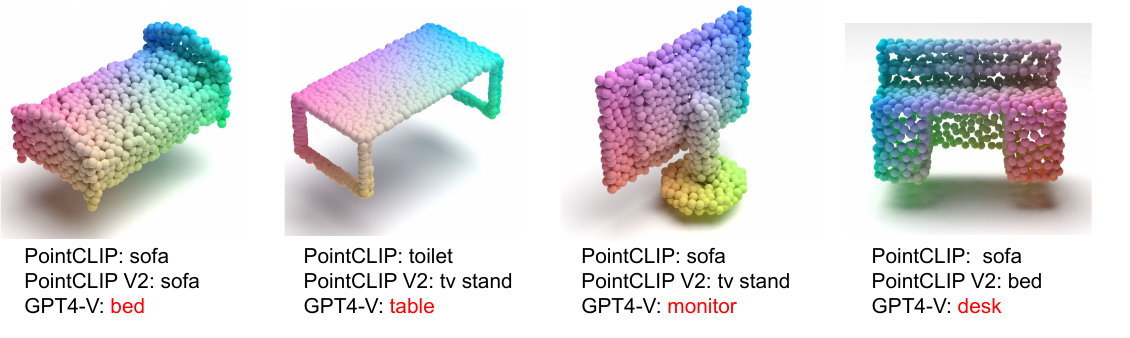}
    \caption{\textbf{Qualitative results:} comparison with the-state-of-the-art methods~\citep{zhu2022pointclip, zhang2022pointclip}. GPT-4V makes the right choice while the previous methods fail to do so. Note that the colored image is for point cloud visualization, not for model input.}
    \label{fig:comp-vis}
\end{figure}
\begin{figure}[h]
    \centering
    \includegraphics[width=\linewidth]{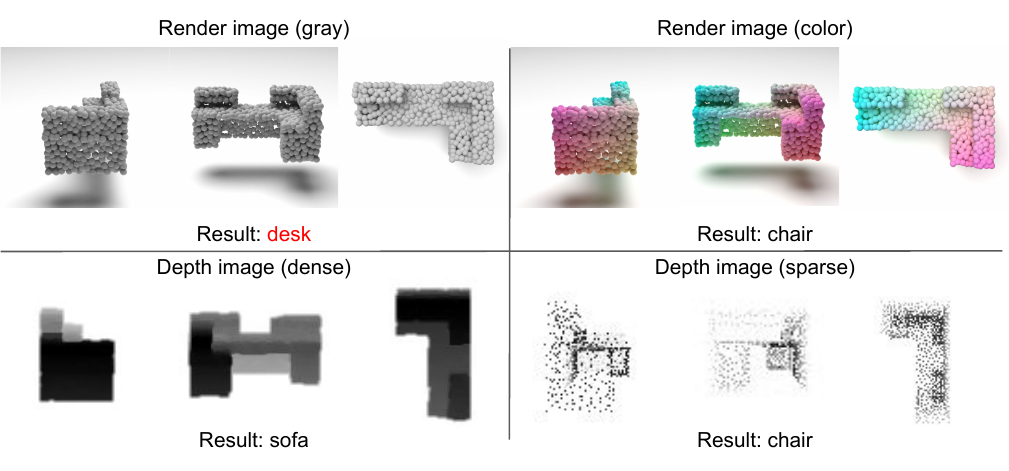}
    \caption{\textbf{Qualitative results:} comparison with the influence of different visualization methods on the results of GPT-4V decision. Among four visualizations, only rendered gray image helps GPT-4V make the right classification.}
    \label{fig:abl-vis}
\end{figure}

As illustrated in Figure~\ref{fig:comp-vis}, GPT-4V accurately classifies the target image into its correct category, outperforming PointCLIP and PointCLIP V2, which fail in this task. Further, Figure~\ref{fig:abl-vis} demonstrates that the image rendered in gray shades offers the most realistic representation among the tested visualizations. This rendering approach more effectively captures the true geometry and distinctive features of the desk, compared to the depth maps, which provide a less detailed depiction.

\begin{figure}[h]
    \centering
    \includegraphics[width=\linewidth]{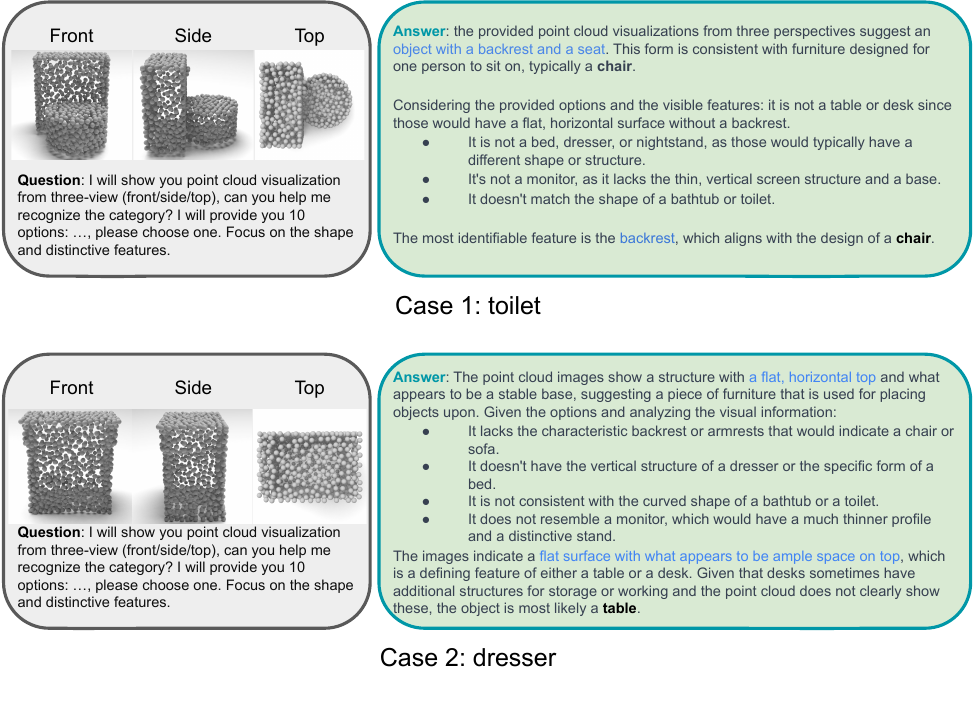}
    \caption{\textbf{Frequent failure cases for GPT-4V.} Case 1: GPT-4V identifies the backreset as the mean feature of chair, neglecting the possibility to be toilet tank. Case 2: GPT-4V can hardly distinguish dresser, night stand or table with the nearly rectangular cuboid point cloud alone.}
    \label{fig:fail}
\end{figure}

\revise{\section{Limitations}
}
\revise{\subsection{When GPT-4V fails?}}
Most of failure cases can fall into two categories: 
(1) \textbf{overconfident on one object feature}. As illustrated by Figure~\ref{fig:fail} case1, GPT-4V emphasize the existence of backrest thus identifying the object as chair, but ignoring the possibility of being a toilet tank. 
(2) \textbf{Less information provided by point cloud}. In the second case, the provided point cloud bring the ambiguity for identifying the true category, especially with the approximately rectangular cuboid shape. It is hard to distinguish whether the demonstrated point cloud blongs to table, night stand or dresser, even for human evaluators. Being lack of surface and texture, GPT-4V itself explained that ``it can be challenging to distinguish further, as point cloud visualizations can lack the finer details necessary for a more definitive identification".
\revise{\subsection{Inference time}

Despite the robustness and high accuracy of GPT-4V in understanding point clouds, a notable limitation is its slow inference speed. This is primarily attributed to its autoregressive generative architecture\footnote{The inference time of GPT-4V also depends on the network latency, so we take time in five cases in average. }  and substantial model size, which contrasts with CLIP-based methods~\citep{zhu2022pointclip, zhang2022pointclip} that only require a feed-forward pass. A qualitative comparison of inference times is provided in Table~\ref{tab:time}, highlighting this efficiency gap.}
\begin{table}[]
    \centering
    \caption{\color{black}Quantitative comparison on inference time with CLIP-based methods.}
    \resizebox{0.5\linewidth}{!}
    {
    \color{black} 
    \begin{tabular}{l|ccc}
        Methods  & PointCLIP & PointCLIP V2  & Ours (3) \\
        \hline
          Inference time(s) & 0.167 & 0.0471 & 5.02
        \vspace{-8mm}
    \end{tabular}}
    \label{tab:time}
\end{table}

\section{Details about datasets}
\textbf{ModelNet40}~\citep{uy2019revisiting} is a the most widely adopted benchmark for point-cloud classification. It contains objects from 40 common categories. There are 9840 objects in the training set and 2468 in the test set. Objects are aligned to a common up and front direction.

\textbf{ModelNet10} is a part of ModelNet40 dataset, containing 4899 pre-aligned shapes from 10 categories. There are 3991 (80$\%$) shapes for training and 908 (20$\%$) shapes for testing.


\end{document}